%% file: paper_ocsr_markush.tex
\documentclass{article}

\usepackage{natbib}
\usepackage{edisonstyle}

\usepackage{amsfonts}
\usepackage{nicefrac}
\usepackage{microtype}
\usepackage{subcaption}
\usepackage{siunitx}
\usepackage{threeparttable}
\usepackage{multirow}
\usepackage{placeins}
\usepackage{enumitem}
\usepackage{xspace}

\hypersetup{hidelinks} 
\newcommand{\methodbase}{MarkushGlyph}
\newcommand{\method}{\methodbase\xspace}
\newcommand{\ocsrmethodbase}{OCSRGlyph}
\newcommand{\ocsrmethod}{\ocsrmethodbase\xspace}

\newcolumntype{Y}{>{\raggedright\arraybackslash}X}
\edisontitle{MarkushGlyph and OCSRGlyph: Improved Chemical Structure Recognition}
\edisonauthors{Alex Andonian$^{1}$, Samuel G Rodriques$^{1}$, Andrew D White$^{1}$, Siddharth Narayanan$^{1}$}
\edisonaffil{$^{1}$Edison Scientific \\ Correspondence to \texttt{sid@edisonscientific.com}}
\edisonabstract{\input{sections/ocsr_markush/abstract}}
\edisonlogo{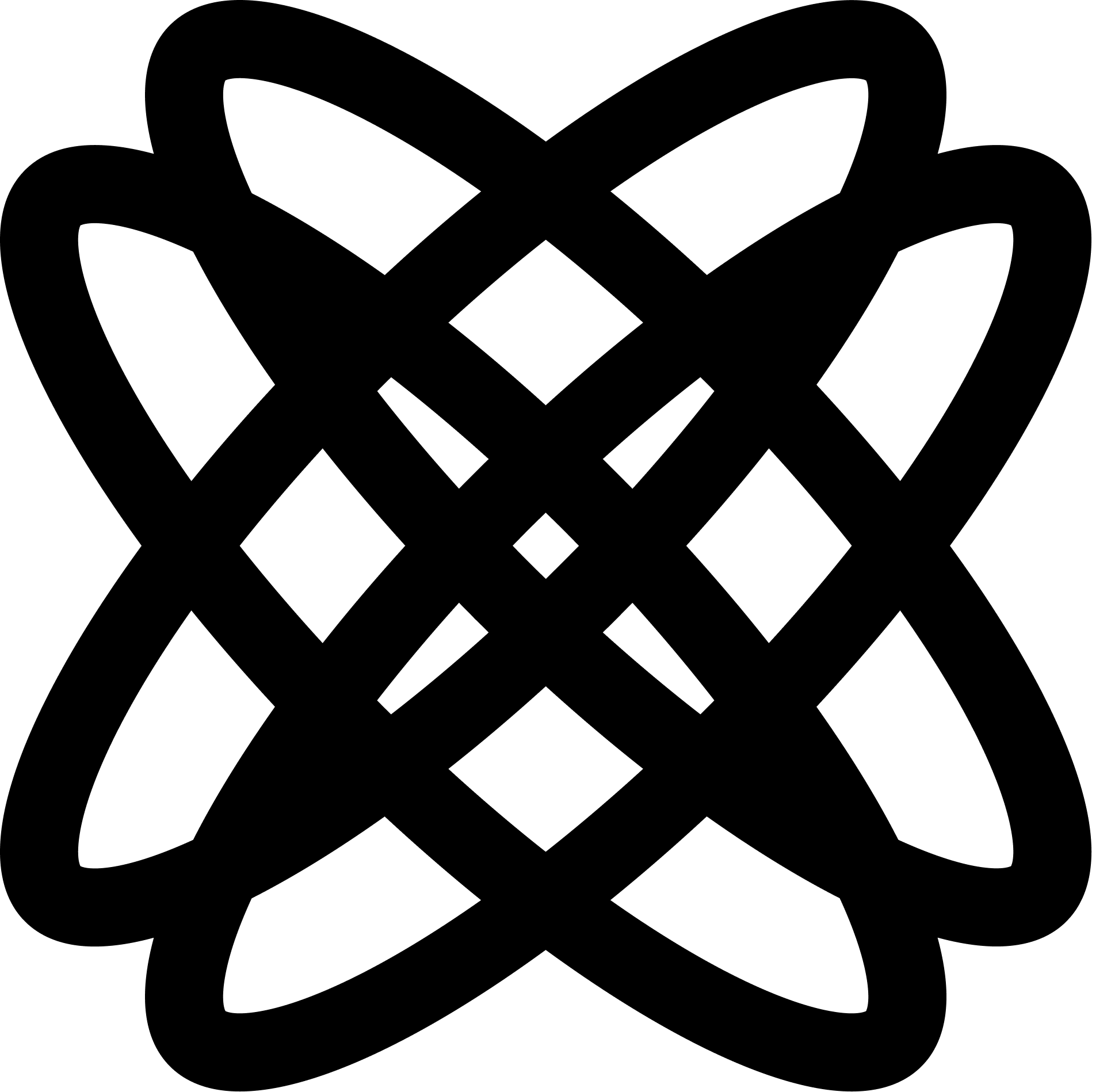}

\begin{document}

\makeedisonheader

\input{sections/ocsr_markush/introduction}
\input{sections/ocsr_markush/related_work}
\input{sections/ocsr_markush/method}
\input{sections/ocsr_markush/datasets}
\input{sections/ocsr_markush/experiments}
\input{sections/ocsr_markush/conclusion}

\clearpage
\appendix
\input{sections/ocsr_markush/appendix_a}

\input{sections/ocsr_markush/appendix_b}

\bibliographystyle{abbrvnat}
\bibliography{references}

\end{document}

%% file: sections/ocsr_markush/abstract.tex
Chemical structures appear in patents and the scientific literature as images. For programmatic
usage, such as indexing in databases or constructing machine learning model training sets, they
must be transformed into line notations. The two common forms of this task are translating
an image of a single molecule (optical chemical structure recognition - OCSR) and translating a Markush 
structure that represents a family of molecules. While prior work in the former case is quite
mature, Markush structure parsing remains a challenging task. In this work, we treat both tasks
as an image-to-text translation problem. We then propose \ocsrmethod, a state-of-the-art OCSR model,
improving performance over prior methods by carefully considering stereochemistry. For the Markush
task, we introduce \method, a vision-language model that reads the entire Markush structure as an 
image. This contrasts with prior systems, which often use multiple stages to separately process
visual and text input content. Finally, we introduce a new metric for determining the accuracy
of Markush structure translations, handling failure modes present in prior metrics.

%% file: sections/ocsr_markush/introduction.tex
\section{Introduction}
\label{sec:introduction}

\begin{figure}[tbp]
    \centering
    \includegraphics[width=\linewidth]{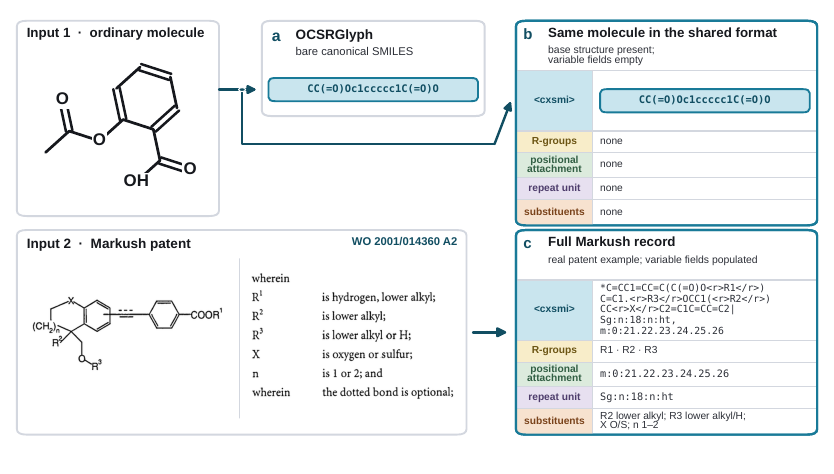}
    \caption{From a single molecule to a family of related compounds.
    Ordinary optical chemical structure recognition converts an image into a text representation of one molecule. Markush structure translation instead outputs a more general representation that encodes a common scaffold together with variable sites and constraints that describe many
    related compounds. Both tasks require reading the same atoms and bonds from the drawing. In the structured
    target format used here (CXSMILES), an ordinary molecule has no variable-group fields, while a
    Markush record adds R-groups, positional attachment information, repeat
    units, and substituent definitions.}
    \label{fig:task_scope}
\end{figure}

Much of chemistry is communicated as images. A structure drawn in a patent or a
journal article is immediately readable to a chemist but cannot be easily parsed for programmatic use. Recovering the underlying structure
from such an image is a prerequisite for searching databases by structure,
checking a new compound against prior art, and assembling the large training
sets that downstream chemistry models depend on. 

Turning the drawing of one molecule into a text representation a program can parse is known as optical chemical structure recognition (OCSR). The output is typically a line notation - most commonly the Simplified Molecular Input Line Entry
System (SMILES), which writes a molecular graph as a compact string, listing heavy atoms as
element symbols in a traversal order and marking branches, ring closures, and
bonds other than single bonds, with hydrogen atoms usually left implicit \citep{weininger1988smiles}.
Modern OCSR systems read ordinary molecules from images with high accuracy
\citep{rajan2021decimer,qian2023molscribe,xu2024molnextr,fang2025molparser}. 

A Markush structure is a more general visual molecular representation used in patents. Rather than a
single molecule, it describes a family of related compounds. 
This family is defined by a shared scaffold in the diagram, and one or more further elements: a set of labeled variable sites such as ``R1'' and ``R2'', an optional block of text stating what each site is allowed to be, annotations of group attachment positions, and repeated substructures.
Markush structures are the working vocabulary of patent claims, yet comprehensive
indexing of them still lives almost entirely in proprietary collections
\citep{simmons2003markush}.

The two translation tasks share similarities: they both require interpreting a 
drawing, identifying its atoms, bonds, and labels together with the spatial
relationships among them, and expressing the result in line notation. 
Markush structure translation also requires the resolution of variable sites and 
constraints. In a sense, OCSR can be viewed as a special-case of the Markush 
translation task, without any variable groups or substituents~
(\autoref{fig:task_scope}). In particular, it is important to note that the
standard output representation for Markush structure translation (CXSMILES) can
also represent individual molecules.

In practice, Markush translation poses a few additional difficulties. Drawing 
conventions shift across patent offices and over time, resulting in a spectrum
of visual dialects that must be handled. Text defining variable sites can be
written in prose, sometimes conditionally, and sometimes by reference to other
definitions. Finally, there is a paucity of publicly-available data, relative to
OCSR.

Prior work approaches these challenges by employing multiple chemistry training
phases (e.g. \citep{zhang2026molsight}) or staged pipelines to separately parse 
text, layout, and images (e.g. \citep{strohmeyer2026markushgrapher2}). In this work,
we take a different approach: for each task, we train a single model with a single
chemistry training phase, instead relying on general image and text pretraining.
The work is organized around four choices that also serve as its contributions.

\textbf{OCSR accuracy benefits from training data curation.}
Like prior work, we find that recognizing stereogenic elements poses a special
challenge in OCSR. We address this by oversampling chiral molecules in our
training dataset, improving stereochemistry performance without degrading
accuracy on non-chiral molecules. This results in \ocsrmethod, a transformer
encoder-decoder with state-of-the-art exact-match accuracy of 93.8\% on the USPTO 
benchmark of 5{,}719 molecules from United States patents \citep{heidenreich2024uspto}.

\textbf{A stricter Markush structure translation evaluation.}
We demonstrate that a standard metric for Markush structure translation 
(\cite{strohmeyer2026markushgrapher2}) has a failure mode: it faithfully checks
that the required molecular features are present, but it does not always penalize
translations that include additional, incorrect features. We introduce a stricter
test: strict parsed-graph equality. It marks two CXSMILES strings as equivalent if and only
if an atom mapping can be found between the two structures that preserves the
molecular graph, R-group assignments, attachment positions, repeating-unit markings, 
and disconnected fragments. We report both metrics for all evaluable methods.

\textbf{A vision-language model outperforms multi-stage pipelines.}
\method is based on a pre-trained VLM. It accepts the entire structure image and
surrounding text as a single image via the vision encoder, and decodes the translated
output token-by-token. This simpler method results in state-of-the-art performance
across all three standard Markush translation benchmarks: IP5-M, M2S, and USPTO-M.

\textbf{Combined training.}
Because OCSR can be treated as a special case of Markush structure translation, 
we incorporate both tasks into the \method training objective. This enables 
\method to be used
in parsing patents and scientific literature without \emph{a priori} knowledge
of what type of chemical structure is being parsed.

%% file: sections/ocsr_markush/related_work.tex
\section{Related Work}
\label{sec:related_work}

\paragraph{Standard OCSR.}
OCSR systems fall into two families, one that decodes a notation string
directly from the image and one that first reconstructs an explicit molecular
graph. The direct approach treats recognition as image-to-sequence
generation. DECIMER established this line with a Transformer decoder that emits
SMILES \citep{rajan2021decimer}, and DECIMER.ai carried it into the harder
setting of structures embedded in scientific documents
\citep{rajan2023decimerai}. The graph-reconstruction approach instead predicts
atoms, their coordinates, and the bonds between them, and leaves a
cheminformatics toolkit to assemble the final graph and reduce it to a single
standard form. MolScribe, MolGrapher, and MolNexTR are representative
\citep{qian2023molscribe,morin2023molgrapher,xu2024molnextr}. Recent systems
push accuracy further with larger image-text corpora and vision-language
backbones, which process an image and text together, including MolParser and
MolSight \citep{fang2025molparser,zhang2026molsight}, while MolSeek-OCR adapts
a general optical-recognition model to chemical drawings
\citep{tang2026molseekocr}. These systems are usually ranked on shared OCSR
benchmarks, most prominently one built from United States Patent and Trademark
Office (USPTO) filings \citep{heidenreich2024uspto}. Reported accuracies remain
challenging to compare across families, because two systems can share a
benchmark yet differ in their canonicalizers, their example filters, their
handling of invalid references, and their treatment of stereochemistry.

\paragraph{Markush recognition.}
Systems that recognize generic (Markush) structures have largely inherited the
image-only paradigm of optical chemical structure recognition, treating variable
groups as in-image abbreviations without consuming the accompanying
substituent-definition text
\citep{qian2023molscribe,xu2024molnextr,rajan2021decimer,fang2025molparser}. MarkushGrapher was among the first to fuse the two, combining molecular
imagery with the surrounding text and layout, and reported results on the M2S
benchmark \citep{morin2025markushgrapher}. MarkushGrapher-2 extended this into
a staged pipeline whose components read text, layout, and molecular imagery
separately before a fusion stage produces the structured prediction
\citep{strohmeyer2026markushgrapher2}. Alongside the model, it released three
evaluation benchmarks, IP5-M, M2S, and USPTO-Markush. The three draw patent
structures from different sources and differ substantially in size, and
together they form the main public reference for Markush recognition. A
meaningful comparison against them still requires the same examples and a
single fixed scoring procedure for every system, a point we return to in
Section~\ref{sec:experiments}.

\paragraph{Chemical representations.}
Both lines of work depend on how a molecular structure and its annotations are
serialized. A single molecule can be written as a SMILES string in many
equivalent ways, one for each order in which its atoms are visited
\citep{weininger1988smiles,opensmiles}. Canonicalization removes that ambiguity
by choosing one reproducible ordering, so that two spellings of the same
molecule compare as equal. Markush structures need more than the base graph can
express, and extended formats supply it. CXSMILES, the ChemAxon extension of
SMILES, attaches information such as variable-group labels, disconnected
fragments, and repeating or multicenter units to specific atoms by index
\citep{chemaxon_cxsmiles}. This
expressivity carries a cost for recognition. When the atom order changes, a
system must keep every atom-indexed annotation pointing at the right atom, so
parsing and canonicalization become part of the recognition problem rather than
a downstream afterthought. The format used in this work, \texttt{cxsmiles\_opt},
is MarkushGrapher-2's optimized CXSMILES, which moves each label inline beside
the atom it modifies so that labels stay attached when the atom order changes
\citep{strohmeyer2026markushgrapher2}.

\paragraph{Evaluation and adaptation.}
How recognition is scored matters as much as how it is performed. Similarity
and recall-style metrics can award partial credit,
which is misleading when an application needs the exact structure, as patent
novelty assessment does. Earlier OCSR work made this case and argued for
comparing full connectivity tables rather than approximate similarity
\citep{krasnov2024ocsr}. Testing whether two parsed structures are truly
identical is a graph isomorphism problem, that is, deciding whether one
consistent matching of atoms aligns both graphs edge for edge. Standard tools
address it, including VF2-style search over graphs represented in libraries
such as NetworkX \citep{cordella2001vf2,hagberg2008networkx}. On the modeling
side, \method builds on a compact open vision-language
model from the Qwen family, which reads an image and text jointly in a single
network, an arrangement often called early fusion
\citep{qwen2025qwen3,bai2025qwen25vl}. Adapting such a model to a new task need
not retrain all of its parameters. Low-rank adaptation (LoRA) instead learns a
small set of low-rank weight updates, which specializes the model at a fraction
of the cost \citep{hu2021lora}.

%% file: sections/ocsr_markush/method.tex
\section{Method}
\label{sec:method}

We frame OCSR and Markush structure recognition as image-to-sequence problems.
The input is a cropped chemical drawing together with any accompanying printed
text, and the output is a machine-readable structure. For an ordinary
molecule the structure is a single molecular graph. For a Markush drawing it is
a family of graphs, described by a common scaffold, labeled variable sites, and
the constraints among them. We study this formulation with two models. \ocsrmethod reads single molecules, and \method reads either kind of input and produces one structured record that covers both.

\subsection{Target representation}
\label{sec:target_representation}

\method emits a record with two parts, a structure string and the printed
substituent-definition text. An example from the IP5-M benchmark reads
\begin{quote}
\footnotesize
\verb!<markush><cxsmi>CN1C=CN=N1.C<r>Rv</r>|Sg:n:7:r:ht,m:6:1.2.3.4.5</cxsmi>!\\
\verb!<stable></stable></markush>!
\end{quote}
The \texttt{<cxsmi></cxsmi>} tag contains the core chemical structure and Markush annotations.
Read left to right, the example contains two dot-separated fragments, a methyl-substituted five-membered ring
and a carbon bearing the variable group Rv. The label is written inline at its
attachment atom as \verb|<r>Rv</r>|. Two annotation sections follow the
base structure string. The repeat-unit section (\texttt{Sg:}) marks the atoms
of a repeating unit together with its printed repeat annotation. The
positional-variation section (\texttt{m:}) records the ring atoms the labeled
group may bond to, here any one of the five listed.
Disconnected fragments such as counterions remain dot-separated components.
The \texttt{<stable></stable>} tag holds any substituent definitions printed
with the drawing, and is empty in this example. A drawing that defines its
substituents places them here, for example \verb|<stable>R3:CH3</stable>|.

This target format is an optimized form of CXSMILES introduced with
MarkushGrapher-2, which we refer to as \texttt{cxsmiles\_opt}
\citep{strohmeyer2026markushgrapher2}. Standard CXSMILES stores every label in
an atom-indexed table at the end of the string. In that form the example above
becomes
\begin{quote}
\footnotesize
\verb!CN1C=CN=N1.C[*] |$;;;;;;;Rv$,Sg:n:7:r:ht,m:6:1.2.3.4.5|!
\end{quote}
with the label Rv held in the trailing \verb|$...$| table and a placeholder atom
\verb|[*]| in the structure. \texttt{cxsmiles\_opt} moves each label inline
beside the atom it modifies, so the label does not have to be re-indexed when the
atom order changes, while the \texttt{m:} and \texttt{Sg:} sections keep their
annotation form.

An ordinary molecule uses the same record with no variable-site annotations and
an empty substituent definition table, so it is the special case in which the family
collapses to a single molecule. \method therefore emits \texttt{cxsmiles\_opt}
for both Markush structures and ordinary molecules, using a single output format
across the two tasks.

During evaluation, \texttt{cxsmiles\_opt} predictions and their reference targets are canonicalized before
comparison, so that a difference in atom order or an equivalent serialization of
the same structure never counts as an error. Canonicalization fixes one atom
order for the molecular graph and rewrites the annotations against that order,
keeping label identity, attachment sets, repeat-unit membership, and fragment
structure intact. Appendix~\ref{app:eval_record} gives the full serialization and
its edge cases.

\subsection{\ocsrmethod}
\label{sec:ocsr_model}

\ocsrmethod is a standard image-to-sequence architecture. A Swin-B
Transformer encoder at $384$-pixel resolution, initialized from
ImageNet-pretrained weights, reads the image \citep{deng2009imagenet,liu2021swin}.
A six-layer Transformer decoder, trained from scratch, generates a SMILES string
one character at a time through a character vocabulary of $101$ tokens. This is a
bare SMILES string, not the unified \texttt{cxsmiles\_opt} record of
Section~\ref{sec:target_representation}. \ocsrmethod is specialized for ordinary
molecules and does not use that record or its empty-field special case.
The encoder and decoder together total about 94 million parameters, roughly 20
times smaller than the 2B-parameter \method. Decoding is greedy. A deterministic postprocess removes isolated explicit-hydrogen
fragments that the decoder occasionally hallucinates, then re-canonicalizes the
prediction with RDKit \citep{rdkit}.

\subsection{\method}
\label{sec:markush_model}

\method is a vision-language model that generates the structured record
directly from the image. It is built on \texttt{Qwen3.5-2B-Base}, which uses a standard ViT 
vision encoder that feeds representations into a Transformer decoder; the cropped image and a 
short text prompt enter the same network and are processed together \citep{qwen35_2b_base}. 
The printed atom labels, R-group labels, brackets, and substituent text remain in the image
and are read by the model's vision encoder rather than supplied through a
separate text input, so there is no upstream text-recognition stage. The
standard use case is translating Markush structures in patent PDFs, where reading
the printed text directly from the image avoids a separate PDF text-extraction
step. We tune the base model with low-rank adaptation (LoRA) at rank $128$ and scale
$128$ \citep{hu2021lora} in a supervised fine-tuning stage.

\begin{figure}[tbp]
    \centering
    \includegraphics[width=0.95\linewidth]{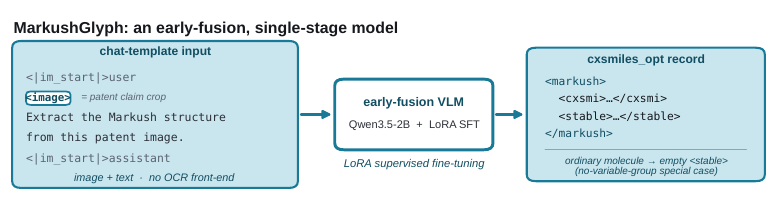}
    \caption{Architecture of \method. A cropped image and a short prompt
    are passed to \texttt{Qwen3.5-2B-Base}, which emits the structured
    \texttt{cxsmiles\_opt} record directly after a LoRA supervised fine-tuning
    stage. An ordinary molecule is the special case whose substituent extension
    is empty.}
    \label{fig:markush_pipeline}
\end{figure}

We use two inference settings. Greedy decoding is used in the default setting.
When employing majority voting, we sample eight candidates with
temperature $0.7$ and nucleus sampling at $0.95$, map each valid candidate to
its canonical parsed-graph form, and return the most frequent one. Both settings
produce the same kind of record.
Appendix~\ref{app:release} gives the full decoding configuration, and the
reproduction commands are in the public code release. Figure~\ref{fig:markush_pipeline}
summarizes this architecture.

%% file: sections/ocsr_markush/datasets.tex
\section{Data and Evaluation}
\label{sec:datasets}

We train two models on separate data and evaluate them on benchmarks that
are held out from training.

\begin{table}[tb]
\centering
\small
\setlength{\tabcolsep}{4pt}
\begin{threeparttable}
\caption{Training data for \ocsrmethod\ and \method. For \ocsrmethod,
\emph{Repeat} is how many times the loader draws each record per epoch, so
\emph{Effective} $=$ \emph{Molecules} $\times$ \emph{Repeat} is the number of records seen. For
\method, \emph{Share} is each source's percentage of the $250{,}000$-record
Markush training set. The ordinary-molecule row is an add-on rather than part of that set:
its \texttt{+5} is its $5\%$ share of the enlarged $263{,}158$-record total
($13{,}158$ of $263{,}158$).}
\label{tab:training_data}
\begin{tabular*}{\textwidth}{@{\extracolsep{\fill}}
    l l S[table-format=7.0] c S[table-format=7.0]
    !{\hspace{1.5em}}
    l l S[table-format=6.0] S[table-format=+2.0] @{}}
\toprule
\multicolumn{5}{c}{\textbf{\ocsrmethod} (image $\to$ SMILES)} &
\multicolumn{4}{c}{\textbf{\method} (image $\to$ record)} \\
\cmidrule(lr){1-5}\cmidrule(lr){6-9}
Source & Role & {Molecules} & Repeat & {Effective} &
Source & Type & {Records} & {Share (\%)} \\
\midrule
PubChem-1M\tnote{a}    & base       & 1000000 & $\times$1 & 1000000 &
  Synthetic\tnote{d}   & synthetic  & 152620  & 61 \\
USPTO-680K\tnote{a}    & base       & 680220  & $\times$1 & 680220  &
  Structure-only\tnote{e} & real    & 52524   & 21 \\
Stereo-200K\tnote{b}   & enrichment & 200000  & $\times$2 & 400000  &
  Molecule$\to$Markush\tnote{f} & real & 44856 & 18 \\
Adjacent-ring\tnote{c} & enrichment & 64752   & $\times$2 & 129504  &
  Ordinary\tnote{g}    & real       & 13158   & +5 \\
\midrule
\bfseries Effective total & & & & \bfseries 2209724 &
  \bfseries Total & & \bfseries 263158 & \\
\bottomrule
\end{tabular*}
\begin{tablenotes}[flushleft,para]
\scriptsize
\item[a] \citealp{xu2024molnextr}
\item[b] \citealp{zhang2026molsight}
\item[c] This work
\item[d] \citealp{morin2025markushgrapher}
\item[e] \citealp{strohmeyer2026markushgrapher2}
\item[f] \citealp{fang2025molparser}
\item[g] \citealp{qian2023molscribe}
\end{tablenotes}
\end{threeparttable}
\end{table}

\subsection{Training data for \ocsrmethod}
\label{sec:ocsr_data}

\ocsrmethod is trained on the image-to-SMILES mixture in
Table~\ref{tab:training_data}. The base pool is the PubChem-1M
\citep{kim2021pubchem} and USPTO-680K OCSR training data distributed together by
MolNexTR. PubChem images are rendered on the fly with
Indigo \citep{indigo}, while the USPTO images are pre-rendered. Targets are
canonical SMILES produced by RDKit \citep{rdkit}. The USPTO split is disjoint
from the USPTO evaluation benchmark.

Analysis of early training runs traced the residual errors to stereochemistry,
where the model recovered the molecular graph but mis-assigned the configuration
at stereocenters. Stereoisomers that share a connectivity graph can be
distinct compounds with distinct properties. We therefore supplement the training
dataset with two stereochemistry-rich sources. One is Stereo-200K 
\citep{zhang2026molsight}, a pre-existing
dataset of stereochemistry-bearing molecule images that we preprocess to a white
background and a tight crop. The other is a $64{,}752$-molecule subset of
PubChem-1M carrying adjacent-ring stereocenters, a motif identified as especially
difficult in early training runs. These molecules are already in PubChem-1M,
so the addition of this subset is equivalent to sampling them twice
as often during training.

\subsection{Training data for \method}
\label{sec:markush_data}

\method is trained on the mixture in Table~\ref{tab:training_data}, where records
are written in the unified \texttt{cxsmiles\_opt} format of
Section~\ref{sec:target_representation}. Three sources make up a $250{,}000$-record
Markush training set, the largest of which consists of synthetic Markush structures
rendered with surrounding patent-image context. The remainder are real Markush
structures from United States patents and real molecules converted into the
record format.

To let \method read ordinary molecules through the same interface, we add
single-molecule examples encoded as the no-variable-group special case of the
record. These are sampled from the USPTO subset used in training \ocsrmethod. 
Our final model is trained on a dataset with a $5\%$ OCSR subset, but we also
report ablations with $10\%$ and $15\%$.

\subsection{Benchmarks}

OCSR is evaluated on the USPTO OCSR benchmark of 5{,}719 images \citep{heidenreich2024uspto},
the held-out set on which the prior
systems we compare against report their accuracies. Three further OCSR evaluation
sets, CLEF-IP (977) \citep{clef_ocsr}, JPO (449) \citep{jpo_ocsr}, and UoB
(5{,}740) \citep{uob_ocsr}, are used only in the analysis of \method's
ordinary-molecule ability.

Markush recognition is evaluated on the three MarkushGrapher-2 benchmark datasets
\citep{strohmeyer2026markushgrapher2,morin2025markushgrapher}, IP5-M (878
examples), M2S (103 examples), and USPTO-Markush (74 examples). Each example is a
Markush structure image derived from a patent. In M2S the image also shows the
printed text that defines the variable groups, and the target includes that text.
IP5-M and USPTO-Markush contain structure images alone, and their targets carry
no substituent-definition text. MarkushGrapher-2 reports IP5-M on a larger set of
1{,}000 examples, but only 878 are publicly released, so we use those 878 as a
fixed denominator for every system in the comparison.

\subsection{Scoring}
\label{sec:experiments_metrics}

The two models are scored differently. \ocsrmethod must emit a bare 
canonical SMILES string. This is scored by exact match using RDKit. There are
multiple stereochemistry conventions used in the literature, which we also
consider here. \emph{Canonical exact match}
requires fully-correct stereochemistry: both tetrahedral chirality and cis/trans
double-bond geometry. The \emph{chirality-kept} score checks tetrahedral chirality but
ignores cis/trans, and finally the \emph{graph} score ignores all stereochemistry. 
We also report the valid-molecule rate. Appendix~\ref{app:eval_ocsr} documents the 
canonicalization steps and per-system conventions in full. Un-parseable outputs are
marked incorrect.

\method emits \texttt{cxsmiles\_opt} for both OCSR and Markush recognition. An OCSR
task instance is scored using the molecular part of the record, but the prediction is
marked incorrect if there are spurious variable-group annotations.
A Markush recognition task instance is scored against the full record.

Markush recognition outputs are scored in two ways. To be consistent with prior work, we first
consider the scoring procedure released with MarkushGrapher-2
\citep{strohmeyer2026markushgrapher2}, using the published implementation.
This scoring function is analogous to a recall measure: it marks a prediction correct when the 
scaffold matches and
the reference's variable-region annotations are present, even when the prediction
adds annotations the reference does not contain or transposes labels on a
symmetric backbone whose annotations break that symmetry
(Figure~\ref{fig:strict_vs_mg2_grading}). It also does not account for stereochemistry.

We further introduce a new metric: strict parsed-graph equality. This scoring function
requires that the entire prediction be correct. It searches for a consistent atom matching
between the proposed and reference structures. This matching must align the molecular 
graph, the R-group assignments, the positional-variation and repeat-unit sections, 
and the fragment correspondence. In this setting, a correct scaffold with a
misplaced or extra annotation fails. Note that neither scoring function checks the
substituent-definition text.

Appendix~\ref{app:eval_detail} details the scorer configuration and the two
Markush metrics. Grader validation and parse-validity counts are in
Appendix~\ref{app:strict_grader}.

%% file: sections/ocsr_markush/experiments.tex
\section{Experiments}
\label{sec:experiments}

We report results on both tasks. OCSR is presented first, then Markush
structure recognition. We then consider multi-task performance and present
a series of ablations to isolate the effects of various design decisions.

\subsection{Main results}
\label{sec:experiments_main}

\paragraph{\ocsrmethod.}
Table~\ref{tab:ocsr_comparison} compares \ocsrmethod with published OCSR systems
on USPTO, where it attains the best accuracy under all three conventions.
Under the strictest convention (canonical exact match), it reaches 93.8\%, with 99.6\% molecule
validity. The prior state of the art, MolSight, is the only baseline to report
both a full-stereo and a graph score, and \ocsrmethod exceeds it by 1.8 percentage points on
canonical exact match and 2.2 on the graph score. Under the chirality-kept
convention that MolScribe, MolParser, and MolNexTR report, \ocsrmethod at 93.9\%
matches the strongest of them: MolNexTR at 93.8\%. 
Appendix~\ref{app:eval_ocsr} gives the full evaluation protocol and
Appendix~\ref{app:ocsr_errors} the error breakdown.

\begin{table}[tb]
\centering
\caption{OCSR accuracy on the USPTO benchmark (5{,}719 images) under three
stereochemistry conventions (\autoref{sec:experiments_metrics}).
Baseline cells are each method's own reported value except those marked
\textsuperscript{m}, which are reported by MolSight.}
\label{tab:ocsr_comparison}
\begin{threeparttable}
\small
\setlength{\tabcolsep}{9pt}
\begin{tabular}{l c c c}
\toprule
\multirow{2}{*}{System} & \multicolumn{3}{c}{Exact-match accuracy (\%)} \\
\cmidrule(lr){2-4}
& Canon. & Chirality-kept & Graph \\
\midrule
DECIMER 2.7~\citep{rajan2021decimer}   & 58.4\tnote{m}  & --            & 61.5\tnote{m} \\
MolGrapher~\citep{morin2023molgrapher} & 65.7\tnote{m}  & --            & 91.5 \\
MolScribe~\citep{qian2023molscribe}    & 88.4\tnote{m}  & 92.6          & 94.6\tnote{m} \\
MolParser~\citep{fang2025molparser}    & --             & 93.0          & -- \\
MolNexTR~\citep{xu2024molnextr}        & --             & 93.8          & -- \\
MolSight~\citep{zhang2026molsight}     & 92.0           & --            & 94.0 \\
\midrule
\method\ (ours)\tnote{\textdagger}     & 69.0           & 72.5          & 81.3 \\
\textbf{\ocsrmethod\ (ours)}           & \textbf{93.8}  & \textbf{93.9} & \textbf{96.2} \\
\bottomrule
\end{tabular}
\begin{tablenotes}[flushleft]
\footnotesize
\item[m] MolSight reports these under a single protocol because the original papers give no value under this convention.
\item[\textdagger] Trained on only $5\%$ ordinary molecules ($13{,}158$; Table~\ref{tab:training_data}).
\end{tablenotes}
\end{threeparttable}
\end{table}

\begin{figure}[tbp]
    \centering
    \includegraphics[width=\linewidth]{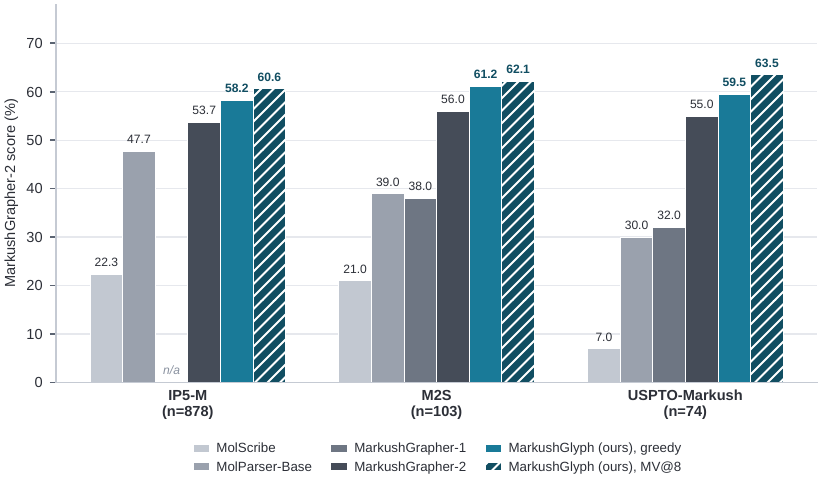}
    \caption{Markush recognition on the benchmarks under the
    MarkushGrapher-2 score. Each group is one benchmark; bars are
    prior systems (gray) and \method (teal) in two decoding settings. \method is above every prior system on all three benchmarks.
    Exact values and the strict parsed-graph results are in
    Table~\ref{tab:markush_headtohead}.}
    \label{fig:markush_bars}
\end{figure}

\paragraph{\method.}
Table~\ref{tab:markush_headtohead} compares \method with other Markush structure
recognition systems on IP5-M ($n=878$), M2S ($n=103$), and USPTO-Markush ($n=74$),
under the MarkushGrapher-2 score and strict parsed-graph equality
(Figure~\ref{fig:markush_bars}). Under both scoring functions, \method exceeds
MarkushGrapher-2 on all three benchmarks in the single-prediction setting. 
Majority voting pushes performance up even further, by 0.9 to 4.0 percentage points.

We only compare to MarkushGrapher-2 under strict parsed-graph equality, as it is
the best prior method, by some margin. To do so, we ran inference using the published model
over the benchmark sets. On IP5-M and M2S, we replicated the reported scores under
the MarkushGrapher-2 metric. However, on USPTO-M, our reproduction only achieved 41.9\%.
We note this is not a typo: in our evaluation, MarkushGrapher-2
achieved the same score on USPTO-M under both metrics.
In \autoref{tab:markush_headtohead}, we report the published numbers 
from~\cite{strohmeyer2026markushgrapher2}, but we note that the strict parsed-graph
equality result on USPTO-M may be affected by this issue.

Strict parsed-graph equality is lower than the MarkushGrapher-2 score in most cases,
because our scoring function typically applies tighter criteria, specifically penalizing incorrect
stereoisomers and additional variable groups. 
Section~\ref{sec:experiments_unification} analyzes the divergence more closely.

\begin{table}[t]
    \centering
    \caption{Markush recognition on the benchmarks under MarkushGrapher-2's
    score and strict parsed-graph equality, both defined in
    Section~\ref{sec:experiments_metrics} (the strict scorer is validated in
    Appendix~\ref{app:strict_grader}). Prior systems' MarkushGrapher-2 scores are the published
    numbers from \citet{strohmeyer2026markushgrapher2}; strict values are scored
    uniformly by our implementation for both systems, MarkushGrapher-2's from its
    released model. Denominators are the full evaluation sets (IP5-M $878$,
    M2S $103$, USPTO-Markush $74$); an unparseable prediction counts as incorrect.
    Bold marks the best value per column.}
    \label{tab:markush_headtohead}
    \begin{threeparttable}
    \small
    \setlength{\tabcolsep}{5pt}
    \begin{tabular}{lrrrrrr}
        \toprule
        & \multicolumn{3}{c}{MarkushGrapher-2 score} & \multicolumn{3}{c}{Strict parsed-graph equality} \\
        \cmidrule(lr){2-4}\cmidrule(lr){5-7}
        Method & IP5-M & M2S & USPTO-M & IP5-M & M2S & USPTO-M \\
        \midrule
        MolScribe \citep{qian2023molscribe} & 22.3 & 21.0 & 7.0 & -- & -- & -- \\
        MolParser-Base \citep{fang2025molparser} & 47.7 & 39.0 & 30.0 & -- & -- & -- \\
        MarkushGrapher-1 \citep{morin2025markushgrapher} & -- & 38.0 & 32.0 & -- & -- & -- \\
        MarkushGrapher-2 \citep{strohmeyer2026markushgrapher2} & 53.7 & 56.0 & 55.0 & 51.1 & 51.5 & 41.9\tnote{\textdagger} \\
        \midrule
        Claude Opus 4.8~\citep{anthropic2026opus48}\tnote{a} & 5.5 & 1.9 & 0.0 & 4.0 & 6.8 & 0.0 \\
        GPT-5.6-sol~\citep{openai2026gpt56}\tnote{a} & 26.3 & 40.8 & 20.3 & 21.2 & 40.8 & 14.9 \\
        \midrule
        \method (ours), single greedy & 58.2 & 61.2 & 59.5 & 52.1 & 61.2 & 54.1 \\
        \method (ours), majority vote ($K=8$) & \textbf{60.6} & \textbf{62.1} & \textbf{63.5} & \textbf{54.7} & \textbf{64.1} & \textbf{56.8} \\
        \bottomrule
    \end{tabular}
    \begin{tablenotes}[flushleft]
    \footnotesize
    \item[\textdagger] Conservative lower bound: our MarkushGrapher-2 reproduction on USPTO-Markush reached $41.9\%$ (published $55.0\%$).
    \item[a] Graded against standard CXSMILES, not \texttt{cxsmiles\_opt}.
    \end{tablenotes}
    \end{threeparttable}
\end{table}

\begin{figure*}[tbp]
    \centering
    \includegraphics[width=\textwidth]{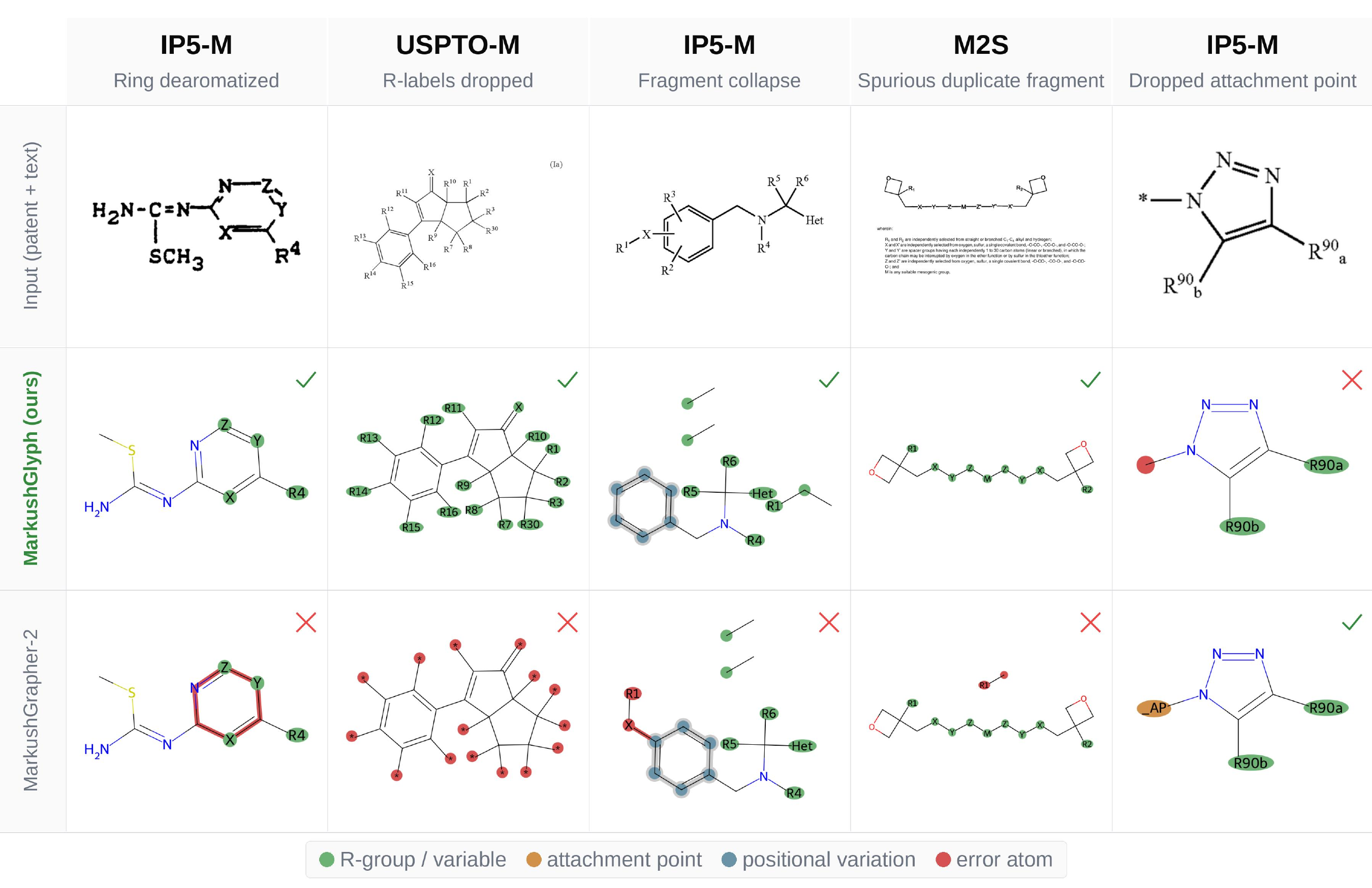}
    \caption{\textbf{Markush qualitative comparison.}
    Five illustrative examples from the Markush benchmarks. Checks and crosses 
    denote pass and fail under the MarkushGrapher-2 scoring function.
    In the first four examples \method passes and MarkushGrapher-2 fails: from left to right,
    MarkushGrapher-2 turns the aromatic ring into a non-aromatic
    single/double-bond pattern (the R4/X/Y/Z labels are unchanged), drops all
    fifteen ring R-positions to generic attachment points, merges the separate R1--X fragment into
    the benzyl ring (four fragments to three), and appends a duplicate R1--C
    fragment. In the fifth, MarkushGrapher-2 passes while \method caps the ring-N
    attachment point with a methyl. MarkushGrapher-2 predictions come from its
    ChemicalOCR pipeline; \method uses majority voting over eight samples.}
    \label{fig:markush_qualitative}
\end{figure*}

\paragraph{Qualitative comparison.}
Figure~\ref{fig:markush_qualitative} shows five illustrative examples from the
benchmark evaluations. In the first four, \method passes the
MarkushGrapher-2 score (the \texttt{cxsmi\_equality} scorer) and MarkushGrapher-2
fails it; the fifth shows the converse.

\FloatBarrier

\paragraph{Frontier vision-language models.}
As a reference point, we run two frontier VLMs, Claude Opus 4.8 and GPT-5.6-sol,
on the same image and instruction \method receives. As reported in 
\autoref{tab:markush_headtohead}, both fall far below the dedicated methods evaluated.
Providing the models with a
short \texttt{cxsmiles\_opt} format specification helps the weaker Claude (to
25.9\% on IP5-M) but slightly hurts GPT-5.6-sol, whose default output is already
standard CXSMILES.

\subsection{Ablations}
\label{sec:experiments_ablations}


\paragraph{OCSR data curation.}
We varied only the training-data curriculum, holding the Swin-B/384 encoder, the
six-layer decoder, the objective, the optimizer, and the 100{,}000-step schedule
fixed, so any change in accuracy is attributable to the training data. Training on
PubChem-1M and USPTO-680K alone plateaus near 90\% canonical exact match.
Oversampling a 50{,}000 stereo-rich subset alone brings performance to 93.5\%, and
the full set described in \autoref{sec:datasets} reaches 93.8\%. 

\begin{figure}[tbp]
    \centering
    \includegraphics[width=\linewidth]{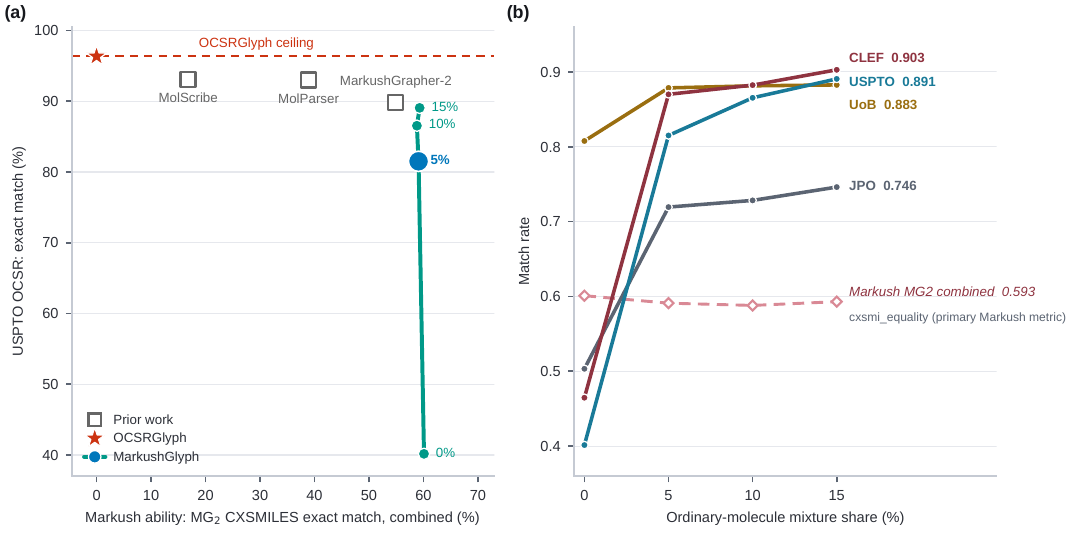}
    \caption{(a)~A capability landscape. Markush ability (horizontal) is the MarkushGrapher-2 combined
    cxsmi\_equality; OCSR ability (vertical) is graph exact match on the USPTO images, one benchmark for every system. As the ordinary-molecule mixture grows from 0 to 15\% (teal
    path), \method climbs the OCSR axis without substantially degrading Markush accuracy. 
    (b)~The same sweep, plotted as accuracy versus mixture share: raising the
    ordinary-molecule share from 0 to 15\% lifts OCSR sharply
    on all four OCSR benchmarks (CLEF, JPO, UoB, USPTO; graph exact
    match).}
    \label{fig:landscape_doseresponse}
\end{figure}

\paragraph{OCSR training in \method.}
In \autoref{fig:landscape_doseresponse}, we illustrate the effect of adding OCSR
samples to the \method training pool. Raising the share of these samples from 0 to 15\%
of the training set leaves Markush structure recognition accuracy nearly flat, 
while raising OCSR performance substantially. 

\subsection{Markush scoring function analysis}
\label{sec:experiments_unification}

\begin{figure}[tbp]
    \centering
    \includegraphics[width=\linewidth]{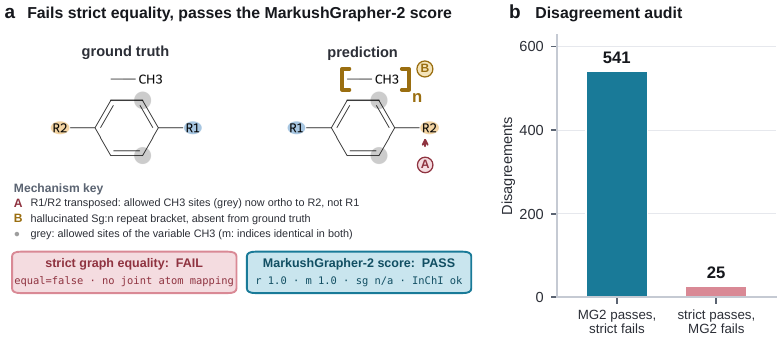}
    \caption{A comparison of the MarkushGrapher-2 and
    strict parsed-graph equality scoring functions. (a)~An incorrect model prediction 
    that passes the former scorer but fails ours. The prediction transposes the R1/R2 labels
    and adds a repeat-unit bracket (\texttt{Sg:n}) over the variable methyl. On
    the symmetric backbone, the MarkushGrapher-2 scorer's per-feature checks do not verify
    the consistency of atom mappings, so it does not catch the mistake.
    Strict equality, however, requires that all features match under one atom bijection
    and therefore fails this prediction.
    (b)~A bulk comparison of agreement over 566 model predictions marked incorrect by only
    one scorer.}
    \label{fig:strict_vs_mg2_grading}
\end{figure}

In \autoref{fig:strict_vs_mg2_grading}, we compare the two scoring methods on an exemplar
prediction. The MarkushGrapher-2 scorer marks this prediction as correct. However, the
prediction transposes the R1/R2 labels and adds a repeat-unit bracket (\texttt{Sg:n}) 
over the variable methyl. We further perform a bulk disagreement analysis, demonstrating
that our scoring function is indeed stricter.

\FloatBarrier

%% file: sections/ocsr_markush/conclusion.tex
\section{Conclusion}
\label{sec:conclusion}

Together, \ocsrmethod and \method reach state-of-the-art accuracy across OCSR
and Markush structure recognition, a step toward a model that translates
both. \ocsrmethod reaches 93.8\% canonical exact match on the USPTO benchmark, and
\method exceeds prior methods on all three Markush benchmarks. Both methods
achieve state-of-the-art performance through relatively simple strategies: leveraging a
pre-trained backbone and a single training phase over carefully-sampled datasets.

We note two shortcomings of this work. First, \method leaves some gap to SOTA performance
on OCSR. Since it may not be known \emph{a priori} whether a figure contains a single
molecule or a Markush structure when bulk-parsing PDFs, unifying SOTA performance
in a single model is desirable. Second, Markush structure recognition performance still
lags substantially behind OCSR. This work demonstrates that simple methods, training at scale,
and inference-time scaling can help close the gap.

%% file: sections/ocsr_markush/appendix_a.tex
\section{Target Format and Metrics in Detail}
\label{app:eval_detail}

The main text describes the metrics only briefly. Here, we specify the metrics and evaluation contract in more detail, including the structured Markush record and its fields, how canonicalization makes two records comparable, the precise definitions of the two Markush equality predicates, and how strict equality uses the full public evaluation dataset with prediction validity reported separately.

\subsection{OCSR target and metrics}
\label{app:eval_ocsr}

Optical chemical structure recognition (OCSR) maps a rendered molecule image to a machine-readable structure string, here the simplified molecular-input line-entry system (SMILES) \citep{weininger1988smiles}.
We evaluate on the United States Patent and Trademark Office (USPTO) OCSR benchmark of 5{,}719 patent images, the set used by the systems we compare against, obtained from the MolScribe release. In this set, 8 images have annotations that cannot be parsed properly by RDKit, and nearly all public evaluation routines in the literature mark these as incorrect. All main-text OCSR results are reported against the full 5{,}719-image set.

Every metric compares RDKit-canonicalized SMILES \citep{rdkit} by string equality:
each string is parsed with \texttt{MolFromSmiles} and re-serialized with
\texttt{MolToSmiles} with its default settings, which give a canonical, isomeric SMILES. Any
model prediction that is not valid SMILES is marked incorrect. The literature does not use a
single stereochemistry convention, so we report three, each matching a convention
used by the systems in Table~\ref{tab:ocsr_comparison}.

\begin{description}[leftmargin=1.2em,itemsep=2pt,topsep=2pt]
  \item[Canonical exact match.] Default RDKit canonicalization with full stereochemistry, ensuring that both tetrahedral chirality and cis/trans double-bond geometry agree. This is the strictest criterion, and the one MolSight reports as its primary metric.
  \item[Chirality-kept exact match.] Tetrahedral chirality required, cis/trans
  ignored, obtained by deleting the \texttt{/} and \texttt{\textbackslash} bond
  marks before the same canonicalization. MolScribe measures accuracy this way, with MolParser and MolNexTR later adopting this convention as well.
  \item[Graph exact match.] All stereochemistry removed with
  \texttt{RemoveStereochemistry} before canonicalizing, so only connectivity is
  compared. Prior work, including MolSight, reports this quantity under the same terminology.
  Despite the name, it is canonical-SMILES equality after stereo removal, not a graph-isomorphism test.
\end{description}

We also report {\bf Valid-SMILES rate}: the fraction of decoded strings that are valid SMILES
strings.


\subsection{The structured Markush record}
\label{app:eval_record}

A Markush structure is a patent claim drawing that defines a family of compounds
through a common scaffold plus variable substituents, so a single image
specifies many molecules at once. Our training target for these images is a structured
Extensible Markup Language (XML) record whose chemical content is a single canonicalized
extended-SMILES (CXSMILES) string. CXSMILES is SMILES augmented with
atom-indexed annotation fields \citep{opensmiles}, allowing one string to carry
both the base molecular graph and the variable-group structure of the claim.

We write this payload in \texttt{cxsmiles\_opt}, an optimized form of
CXSMILES introduced with MarkushGrapher-2
\citep{strohmeyer2026markushgrapher2}. Standard CXSMILES stores every R-group
label in an atom-indexed table at the end of the string and leaves a
placeholder atom in the structure: a carbon bearing the variable group Rv is
written \verb!C[*] |$;Rv$|!, with the label held in the trailing
\verb|$...$| table and matched to its atom by position.
\texttt{cxsmiles\_opt} instead writes the label inline beside the atom it
modifies (\verb|C<r>Rv</r>|, with \verb|<r>_AP</r>| for unnamed attachment
points), so a label never has to be re-indexed when the atom order changes.
The other CXSMILES extensions keep their original annotation form: repeat units and
polymer/SRU S-group brackets use \verb!|Sg:...|! and positional variation
uses \verb!|m:...|!.

Each of these constructs is a field of the record: the base graph, the inline
R-group labels, the positional-variation and repeat-unit sections, and the
dot-separated fragments. A sixth field, the substituent text, is part of the
XML record outside the CXSMILES string. Altogether, we have:

\begin{description}[leftmargin=1.2em,itemsep=2pt,topsep=2pt]
  \item[Base structure.] The fixed molecular skeleton, written as the CXSMILES
  graph.
  \item[R-group labels (R).] The named variable attachment points (R1, R2, \dots)
  bound to specific atoms of the base structure, each representing a set of
 allowed substituents.
  \item[Positional variation (multicenter S-group, m).] A single virtual
  center bonded to a set of candidate atoms, encoding that a fixed group attaches
  at any one of several sites (one of many).
  \item[Polymer / SRU S-group (Sg).] A structural repeating unit bracket over a
  set of atoms carrying a variable repeat count, encoding how many times a
  subunit repeats (frequency variation).
  \item[Fragments.] Disconnected component records that complete the claimed
  structure (for example salt counterions or detached variable pieces).
  \item[Substituent table text.] The free-text substituent constraints from the patent
  (for example ``R1 is a $C_1$--$C_6$ alkyl group''), retained alongside the
  graph.
\end{description}

\paragraph{CXSMILES Canonicalization.}
Before comparison, the chemical content of a record is rewritten into a canonical CXSMILES form that removes differences in atom ordering and notation.
Two records encoding the same base graph and the same R-group, multicenter/positional variation ($m:$), polymer/SRU ($Sg:$), and fragment annotations then parse to comparable graphs regardless of how either string was originally written.
Strict parsed-graph equality requires both sides to canonicalize into a comparable parsed graph, so an example whose reference cannot be canonicalized via RDKit cannot be scored under the strict scoring predicate and is marked incorrect.

\subsection{The two Markush metrics}
\label{app:eval_metrics}

We report two Markush structure recognition metrics. The MarkushGrapher-2 score is previously reported
in the literature, so we retain it for consistency. We further introduce the a strict parsed-graph equality
score that addresses shortcomings in the MarkushGrapher-2 score. Both metrics rely on canonicalized
CXSMILES records.

\begin{description}[leftmargin=1.2em,itemsep=2pt,topsep=2pt]
  \item[MarkushGrapher-2 score.] This is the \texttt{cxsmi\_equality} criterion introduced by
  MarkushGrapher-2 \citep{strohmeyer2026markushgrapher2}. It first matches fragments between
  the model prediction and reference structure, using maximum-common-subgraph (MCS) overlap.
  The matched fragments' backbones are then compared using InChI representations (variable R groups
  are replaced by carbons for this step); note this
  is not sensitive to stereochemistry. The MCS provides an atom correspondence between the two
  structures. This is used to confirm that the model prediction recovers every ground-truth 
  R-group label, multicenter positional-variation ($m$:) section, and polymer/SRU ($Sg$:) section.
  If all fragments are matched and pass these criteria, the prediction is marked correct. Note
  that extra predicted annotations absent from the reference are not penalized.
  
  \item[Strict parsed-graph equality.] This is our stricter proposed metric. We rely on finding
  a graph isomorphism between the model prediction and reference structure. This isomorphism 
  is required to preserve the graph structure, but also node features like R-groups,
  $m$: sections, and $Sg$: sections. Graph isomorphism is checked with VF2 \citep{cordella2001vf2}
  \citep{hagberg2008networkx}. 
\end{description}

\section{OCSR Error Analysis}
\label{app:ocsr_errors}

Here we provide additional OCSR error analysis that motivated our stereochemistry-enriched training curriculum and contributed to state of the art performance.
The full USPTO benchmark has 5{,}719 images; the analysis here uses
USPTO-5704, the parseable subset whose ground truth resolves to a valid
structure (Appendix~\ref{app:eval_ocsr}); an image with no reference structure
cannot be assigned an error class. On USPTO-5704, \ocsrmethod reaches
a canonical exact match of 0.9397 (5360/5704), a graph exact match of 0.9635
(5496/5704), and a valid-SMILES rate of 0.9979 (5692/5704) from a single
image-to-sequence model (Section~\ref{sec:experiments_main}); the corresponding
full-set values on all 5{,}719 images are 93.8\%, 96.2\%, and 99.6\%.

\begin{figure}[tbp]
    \centering
    \includegraphics[width=\linewidth]{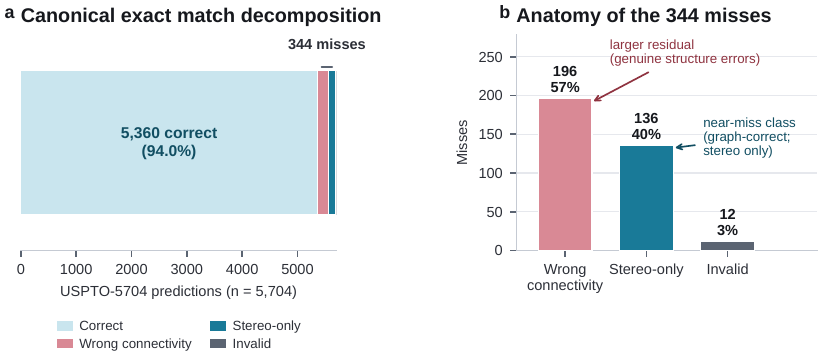}
    \caption{The 344 canonical misses on the USPTO-5704 split into 196
    wrong-connectivity, 136 stereo-only, and 12 invalid-SMILES examples.
    Wrong-connectivity is the largest residual class, and stereochemistry-focused
    curation does not address it; the stereo-only examples are the same molecules
    that the canonical-to-graph metric gap isolates.}
    \label{fig:ocsr_error_anatomy}
\end{figure}

The 344 incorrect predictions on this dataset fall into three classes
(Figure~\ref{fig:ocsr_error_anatomy}): 196 wrong-connectivity errors, 136
stereo-only errors, and 12 invalid SMILES. The stereo-only errors are easily addressable, by inflating the
stereochemistry-enrichment of the training dataset as described in \autoref{sec:ocsr_data}.

%% file: sections/ocsr_markush/appendix_b.tex
\section{Strict parsed-graph equality: validation and examples}
\label{app:strict_grader}

This appendix section opens with a step-by-step description of the strict parsed-graph equality algorithm.
Then, we present additional testing performed to validate the correctness of our implementation,
followed by the error taxonomy that motivated its design and the issues it targeted.
At a high level, our criterion converts two CXSMILES strings into annotated molecular graphs and determines equality on the graph rather than the text, so it accepts equivalent serializations (aromatic versus Kekule ring spelling, different fragment order) and rejects only true differences in connectivity, labels, or annotations.

\paragraph{Algorithm walkthrough.}
The input to the algorithm is a predicted CXSMILES string paired with its ground truth value.
Our evaluator determines equality using the following steps:

(1)~\emph{Parse Base Molecules}: For both the predicted and ground truth CXSMILES, we construct a mapping between atoms in the SMILES base molecule and R group tags, then apply RDKit canonicalization to the SMILES core (preserving explicit hydrogens) and record the atom re-indexing.

(2)~\emph{Parse Markush extensions}: Ingest CXSMILES-specific atom-bond annotations such as \texttt{m:} and \texttt{Sg:}, and reject equality early if any of the above cannot be parsed correctly, or if atom/bond/fragment counts differ.

(3)~\emph{Atom features \& filtering}: Create a feature key for each atom based on its properties (such as its atomic number, formal charge, isotope, etc.), its degree (number of directly bonded atoms), and its neighbor signature, which is a sorted list of (bond features, neighboring atom features) tuples. Bond features indicate a bond's type, stereo configuration, and whether it is aromatic or conjugated.

(4)~\emph{Candidate Mapping}: For ground truth atoms, build a list of predicted atom that have the same feature key as the ground truth. This list of candidates, mapping ground truth atoms to predicted atoms, constrains the full isomorphism search.

(5)~\emph{Isomorphism search}: Apply a graph-isomorphism search to the surviving
pairings, rejecting any partial mapping whose mapped bonds or non-bonds disagree, until it finds a complete bijection.

(6)~\emph{Annotation remap and verdict}: Remap the ground-truth annotation indices through the atom map returned by the isomorphism search, and accept the prediction only if the graph, the remapped annotations, and the resulting structure all agree.

\paragraph{Evaluator validation.}
Before using strict equality to score models, we tested the evaluator itself over 4{,}802 comparisons
(Table~\ref{tab:grading_evidence_appx}, upper block). These comparisons were comprised of 981 unique IP5-M and M2S ground-truth records, each compared against itself, ensuring the evaluator is reflexive, plus the 3{,}821 real ground-truth/prediction pairs captured during bulk evaluation runs. Two checks established correctness. First, every scoring outcoming is reproduced by two independent implementations of graph equality: a NetworkX VF2 isomorphism search
\citep{hagberg2008networkx,cordella2001vf2} and an RDKit canonical-atom-ranking
comparison \citep{rdkit}, which agree with the evaluator on all 4{,}802 comparisons.
Second, all scoring entry points throughout the codebase return an identical
results on every one of the 3{,}821 pairs. Repeating the validation run five times
reproduces the findings; malformed and adversarial predictions (empty strings, truncated XML, invalid SMILES, out-of-range annotation indices) are rejected without crashing or being accepted as equal; the 48 largest structures parse and compare within the search budget; and a deliberately starved budget raises an explicit error rather than silently passing on all three probe cases. 

\paragraph{Error taxonomy of scoring disagreements.}
In order to understand the differences between the MarkushGrapher-2 score and the strict parsed-graph equality score, we compared outcomes on 3{,}821 model-derived predictions from intermediate IP5-M and M2S evaluation runs paired with their ground truth.

On this corpus of predictions, the two scorers agree on 3{,}255 pairs and disagree on 566 (Table~\ref{tab:grading_evidence_appx}, lower block;
Figure~\ref{fig:strict_vs_mg2_grading} in the Experiments section).
These disagreements are highly asymmetric: in 54f1 pairs the MarkushGrapher-2 score credits a prediction that strict equality rejects, whereas the reverse occurs in only 25 pairs.

Interestingly, in all 541 cases, both scorers agree that the predicted base molecule matches the ground truth. Thus, the disagreement arises from incorrectly predicted Markush semantics expressed in a way that the MarkushGrapher-2 score does not penalize.
These errors fall into four classes~(Table~\ref{tab:strict_vs_recall_classes_appx}):
transposed R-group labels on a symmetric backbone, extra labels, a moved or
added positional-variation (\texttt{m:}) section, and an added or wrong polymer/SRU (\texttt{Sg}) annotation, the last two defined in
Appendix~\ref{app:eval_detail}. Transposition errors can be subtle. A swap of two R-group labels should only be credited when the swap occurs across an axis of symmetry with respect to both the backbone and additional Markush annotations.
However, the MarkushGrapher-2 score checks each atom's Markush annotations independently and,
finding both identical, accepts the swap even if it relocates fixed-index \texttt{m:} sites relative
to the R-groups. A transposition onto non-equivalent atoms is rejected by both
scorers. The scorer-verified worked example in
Figure~\ref{fig:strict_vs_mg2_grading} illustrates two of these kinds on a
symmetric benzene backbone: the prediction transposes R1 and R2, which moves the
variable methyl's allowed positions from ortho-to-R1 to ortho-to-R2, and adds a
repeat-unit (\texttt{Sg:n}) bracket the reference does not contain; the
MarkushGrapher-2 score credits it and strict equality rejects it.


\begin{table}[tb]
    \caption{Markush evaluator audit. Upper block: evaluator implementation
    validation, deterministic and stable over generated cases.
    Lower block: A bulk comparison of disagreements over 3{,}821 evaluation pairs}
    \label{tab:grading_evidence_appx}
    \centering
    \footnotesize
    \begin{tabularx}{\linewidth}{@{}l r Y@{}}
        \toprule
        Check & Count & What it establishes \\
        \midrule
        \multicolumn{3}{@{}l}{\textit{Evaluator validation}}\\
        NetworkX VF2 differential & 4{,}802 / 4{,}802 & Isomorphism back-end agrees on every case \\
        RDKit canonical differential & 4{,}802 / 4{,}802 & Independent back-end agrees on every case \\
        Cross-path consistency & 3{,}821 / 3{,}821 & Both code paths return one verdict \\
        Determinism & 4{,}802 / 4{,}802 & Repeated evaluation is identical \\
        Malformed-input fuzz & 10{,}791 / 10{,}791 & Bad input rejected, no false accept \\
        Largest structures & 48 / 48 & Big graphs parse and compare \\
        Timeouts & 3 / 3 & Timeout fires on pathological input \\
        \midrule
        \multicolumn{3}{@{}l}{\textit{Disagreement Comparision (3{,}821 evaluation pairs)}}\\
        Agreements & 3{,}255 & Scorers return the same verdict \\
        Disagreements & 566 & Scorers return different verdicts \\
        \quad MarkushGrapher-2 score true, strict false & 541 & The score credits extra/missing/misplaced semantics \\
        \quad Strict true, MarkushGrapher-2 score false & 25 & Strict accepts a strict-equivalent alternate parsed structure \\
        \bottomrule
    \end{tabularx}
\end{table}

\begin{table}[tb]
    \centering
    \small
    \caption{Error kinds in the 541 cases the MarkushGrapher-2 score credits but strict parsed-graph equality
    rejects. The base molecular overlap is satisfied in every case; the Markush
    semantics differ in a way the MarkushGrapher-2 score does not penalize.}
    \label{tab:strict_vs_recall_classes_appx}
    \begin{tabularx}{\linewidth}{@{}l Y@{}}
        \toprule
        Error kind & Effect that strict equality penalizes \\
        \midrule
        Transposed R-group labels (symmetric backbone) &
            Label names bound to the wrong atoms. On a symmetric backbone the
            per-feature checks pass under inconsistent atom mappings and credit the
            swap; strict requires one fully consistent bijection and rejects it. A swap onto
            non-equivalent atoms is rejected by both scorers. \\
        Extra labels &
            An R-group the target does not contain is added; the prediction's
            apparent scope is broader than the source supports. \\
        Moved multicenter / positional-variation section (\texttt{m:}) &
            The variable attachment shifts to a different atom set, so the
            attachment options differ even when the molecule is identical. \\
        Wrong polymer/SRU S-group annotation (\texttt{Sg}) &
            Repeat or grouping membership changes, so the represented repeat
            scope differs. \\
        \bottomrule
    \end{tabularx}
\end{table}

\FloatBarrier

\FloatBarrier

\section{Reproducibility \& Release}
\label{app:release}

The code and reproduction scripts are released at
\url{https://github.com/EdisonScientific/glyph}. The \ocsrmethod and \method
weights are on Hugging Face at
\url{https://huggingface.co/EdisonScientific/OCSRGlyph} and
\url{https://huggingface.co/EdisonScientific/MarkushGlyph}, and the released
datasets and manifests at
\url{https://huggingface.co/datasets/EdisonScientific/glyph-datasets}.
Table~\ref{tab:release} states the release status of each artifact.
The public benchmarks (IP5-M, M2S, USPTO-Markush) are taken from
\texttt{docling-project/MarkushGrapher-2-Datasets} and are released under
CC-BY-4.0, so they can be redistributed with attribution. Patent-derived images
whose redistribution is restricted are referenced by manifest (dataset
identifiers, row ids, and source pointers) rather than redistributed directly.

\begin{table}[!ht]
    \centering
    \small
    \caption{Release status of paper artifacts.}
    \label{tab:release}
    \begin{tabularx}{\linewidth}{@{}p{0.27\linewidth} l Y@{}}
        \toprule
        Artifact & Status & Notes \\
        \midrule
        Model weights & Released &
            \ocsrmethod and the Markush model (LoRA adapter over Qwen3.5-2B base) on
            Hugging Face as \texttt{EdisonScientific/OCSRGlyph} and
            \texttt{EdisonScientific/MarkushGlyph}. \\
        Training code & Released &
            OCSR scratch-training and Markush LoRA SFT pipelines. \\
        Evaluation code & Released &
            Includes integration with the public MarkushGrapher-2 scorer and the
            strict parsed-graph grader. \\
        Evaluation predictions & Released &
            Per-example predictions and scores supporting the reported results. \\
        Data manifests & Released &
            Example-id manifests for all evaluation sets, with HF config/split
            and source pointers, on \texttt{EdisonScientific/glyph-datasets}. \\
        Evaluation scripts & Released &
            Evaluation and reproduction scripts for the reported protocols. \\
        Public benchmarks (IP5-M / M2S / USPTO-M) & Referenced &
            CC-BY-4.0 via \texttt{docling-project/MarkushGrapher-2-Datasets};
            redistributable with attribution. \\
        Patent-derived images (restricted) & Referenced by manifest &
            Where redistribution is restricted, images are referenced by
            manifest rather than shipped. \\
        \bottomrule
    \end{tabularx}
\end{table}

\FloatBarrier